\documentclass[acmlarge,screen]{acmart}
\setcopyright{none}
\renewcommand\footnotetextcopyrightpermission[1]{}
\usepackage{subfigure}
\usepackage{multirow}
\usepackage{fancyhdr}

\AtBeginDocument{%
  }
\acmJournal{TKDD}
\acmVolume{0}
\acmNumber{0}
\acmYear{xxxx}
\acmArticle{0}

\begin{document}

\title{CSCLog: A Component Subsequence Correlation-Aware Log Anomaly Detection Method}

\author{Ling Chen}
\authornote{The corresponding author.}
\affiliation{%
\department{College of Computer Science and Technology}
  \institution{Zhejiang University}
  \streetaddress{38 Zheda Road}
  \city{Hangzhou}
  \state{Zhejiang}
  \country{China}
  \postcode{310027}
  }
\email{lingchen@cs.zju.edu.cn}

\author{Chaodu Song}
\affiliation{%
  \department{College of Software Technology}
  \institution{Zhejiang University}
  \streetaddress{38 Zheda Rd}
  \city{Hangzhou}
  \country{China}
  \postcode{310027}}
\email{songcd2020@cs.zju.edu.cn}

\author{Xu Wang}

\affiliation{%
  \department{Database Products Business Unit}
  \institution{Alibaba Cloud Intelligence}
  \streetaddress{1008 CaiDent Street}
  \city{Hangzhou}
  \country{China}
  \postcode{310030}}
\email{wx105683@alibaba-inc.com}

\author{Dachao Fu}
\affiliation{%
  \department{Database Products Business Unit}
  \institution{Alibaba Cloud Intelligence}
   \streetaddress{1008 CaiDent Street}
   \city{Hangzhou}
   \country{China}
   \postcode{310030}}
   \email{qianzhen.fdc@alibaba-inc.com}

\author{Feifei Li}
\affiliation{%
  \department{Database Products Business Unit}
  \institution{Alibaba Cloud Intelligence}
   \streetaddress{1008 CaiDent Street}
   \city{Hangzhou}
   \country{China}
   \postcode{310030}}
\email{lifeifei@alibaba-inc.com}

\begin{abstract}
  Anomaly detection based on system logs plays an important role in intelligent operations, which is a challenging task due to the extremely complex log patterns. Existing methods detect anomalies by capturing the sequential dependencies in log sequences, which ignore the interactions of subsequences. To this end, we propose CSCLog, a Component Subsequence Correlation-Aware Log anomaly detection method, which not only captures the sequential dependencies in subsequences, but also models the implicit correlations of subsequences. Specifically, subsequences are extracted from log sequences based on components and the sequential dependencies in subsequences are captured by Long Short-Term Memory Networks (LSTMs). An implicit correlation encoder is introduced to model the implicit correlations of subsequences adaptively. In addition, Graph Convolution Networks (GCNs) are employed to accomplish the information interactions of subsequences. Finally, attention mechanisms are exploited to fuse the embeddings of all subsequences. Extensive experiments on four publicly available log datasets demonstrate the effectiveness of CSCLog, outperforming the best baseline by an average of 7.41\% in Macro F1-Measure.
\end{abstract}





\maketitle

\section{Introduction}
Log data are generated in software systems, e.g., high-performance computing systems, distributed file systems, and cloud services, which are unstructured or semi-structured data with temporal information \cite{zhang2020anomaly}. Log anomaly detection aims to detect system anomalies in a timely manner based on log data, which plays an important role in intelligent operations. As the scale and complexity of the system grow, it becomes more difficult to detect and locate system anomalies manually \cite{liao2013intrusion}. As a result, automated log anomaly detection methods \cite{chen2004failure,liang2007failure,bodik2010fingerprinting,lou2010mining,xu2009detecting,lin2016log,du2017deeplog,zhang2019robust,meng2019loganomaly} are proposed. Generally, in these methods, log messages are parsed into log templates and other features, and detection algorithms are employed to locate anomalies from a series of log messages, i.e., log sequence.

Automated log anomaly detection methods are generally divided into two classes, i.e., traditional methods and neural network-based methods. Traditional methods \cite{chen2004failure,liang2007failure,bodik2010fingerprinting,lou2010mining,xu2009detecting,lin2016log} detect anomalies by extracting the statistical features in log sequences, e.g., number of templates and keyword distribution, and using traditional machine learning models to capture the anomalous information of these features. However, due to the differences in distributions of statistical features of different datasets, it is difficult for these methods to achieve good performance on different datasets. In addition, these methods fail to capture the sequential dependencies in log sequences, which can also indicate the anomalies.

\begin{figure}[!t]
\centerline{\includegraphics[width=0.6\columnwidth]{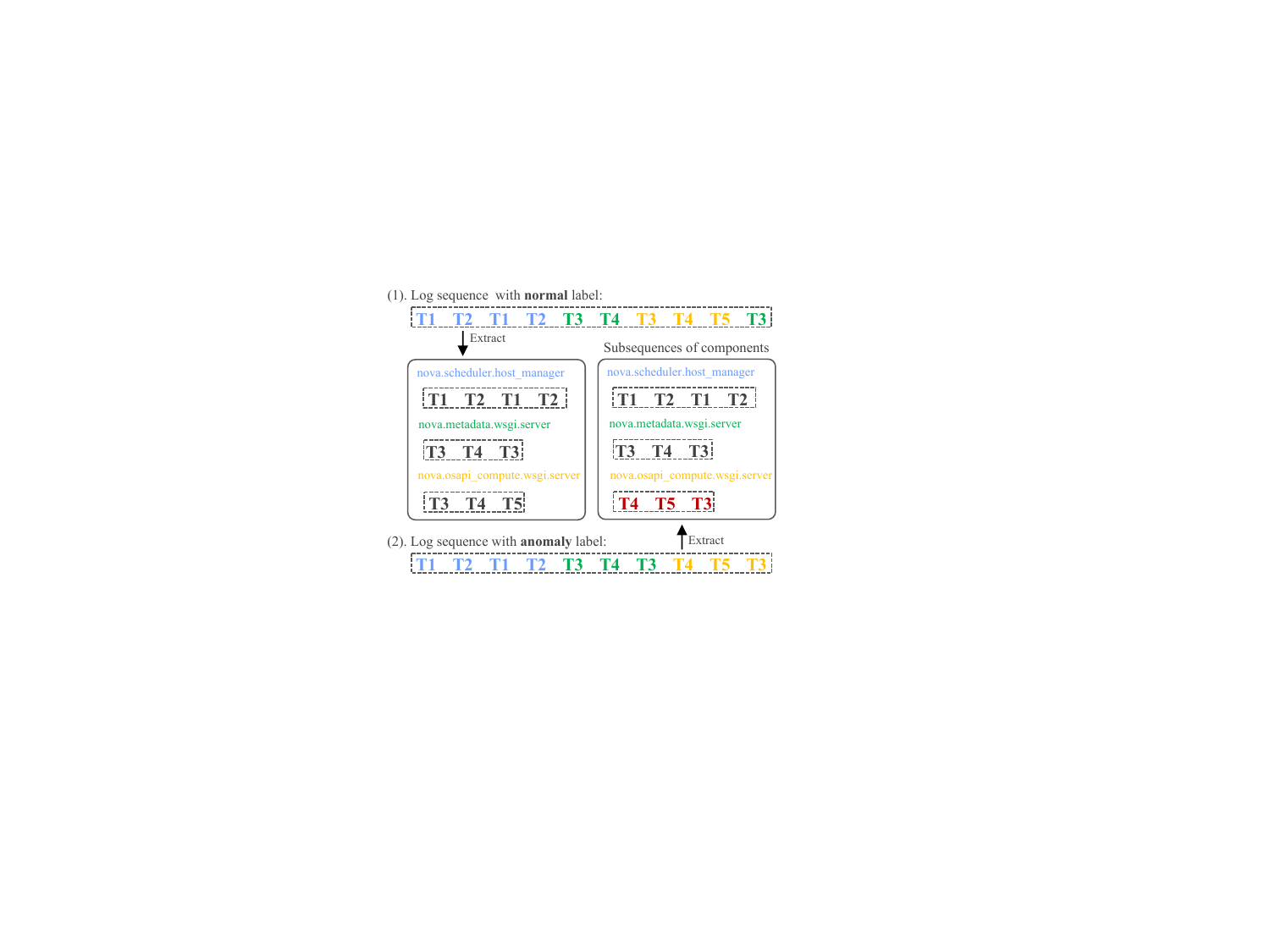}}
\caption{Two log sequences constructed from the OpenStack dataset. The anomalous subsequence of the component ``nova.osapi\_compute.wsgi.server'' is highlighted in red. (Best viewed in color).}
\label{fig1}
\end{figure}

Neural network-based methods \cite{du2017deeplog,zhang2019robust,meng2019loganomaly,li2020swisslog,yin2020improving,nedelkoski2020self,wang2021multi} usually infuse Recurrent Neural Networks (RNNs) and their variants, e.g., Long Short-Term Memory Networks (LSTMs) and Gated Recurrent Units (GRUs), to capture the sequential dependencies in log sequences. These methods focus on mining the patterns of log sequences and modeling them into unified feature representations for anomaly detection. In addition, some methods \cite{han2021unsupervised,xia2021loggan} introduce Generative Adversarial Networks (GANs) to enhance the ability for capturing anomalous information by generating new log data. Some methods \cite{zhang2021logattn,wang2022lightlog} introduce Convolutional Neural Networks (CNNs) to capture the temporal dependencies in log sequences by arranging log templates as a matrix. Some methods \cite{wan2021glad} construct the log sequence as a graph and employ Graph Neural Networks (GNNs) to model information interactions in the graph. However, these methods ignore the interactions of subsequences, which can demonstrate the complex log patterns of components. For example, as shown in Fig. \ref{fig1}, two log sequences on the OpenStack dataset have the same sequence of log templates, but their anomaly labels are different. We find that the subsequences of components extracted from the two log sequences are different. The subsequences of the component ``nova.osapi\_compute.wsgi.server'' are ``T3, T4, T5'' in the normal sequence and ``T4, T5, T3'' in the anomalous sequence. Log template ``T5'' indicates the stop of the compute node in the cloud service, and log template ``T3'' indicates the data requests from the node, which cannot appear after the template ``T5'' in the normal running of the system. This example shows that subsequences can be used to detect the anomaly that is difficult to be detected only by capturing the sequential dependencies in log sequences.

To address the aforementioned problems, we propose CSCLog, a Component Subsequence Correlation-Aware Log anomaly detection method, which not only captures the sequential dependencies in subsequences, but also models the implicit correlations of subsequences. Specifically, the main contributions are outlined as follows:
\begin{itemize}
\item We introduce a subsequence modeling module to extract subsequences from log sequences based on components, and exploit LSTMs to capture the sequential dependencies in subsequences, which can model the complex log patterns of components.
\item We introduce an implicit correlation encoder to model the implicit correlations of subsequences adaptively, and employ Graph Convolutional Networks (GCNs) to accomplish the information interactions of subsequences, which can model the influences between subsequences.
\item We conduct extensive experiments on four publicly available log datasets. Experimental results demonstrate the state-of-the-art (SOTA) performances of CSCLog, outperforming the best baseline by an average of 7.41\% in Macro F1-Measure.
\end{itemize}

\section{Related Work}
In this section, we provide an overview of works related to log anomaly detection, including traditional methods and neural network-based methods.

\subsection{Traditional Methods}
Traditional methods for log anomaly detection can be generally divided into two classes, i.e., supervised learning-based traditional methods and unsupervised or self-supervised learning-based traditional methods.

Supervised learning-based traditional methods aim to classify log sequences by extracting statistical features, e.g., the number of templates and keyword distribution, which require the data with labels. Chen et al. \cite{chen2004failure} introduced decision trees, which have strong interpretability, to model the number of templates in log sequences. Liang et al. \cite{liang2007failure} extracted six different features from log sequences, e.g., time interval and the number of templates, as the inputs of their classifier. Recently, Bodik et al. \cite{bodik2010fingerprinting} employed the Support Vector Machine (SVM) based on the Gaussian kernel to model features in log messages, e.g., frequency of occurrence, periodicity, and correlation of anomalies. Since the frequency of anomalies in a system is low and it is difficult to get a large amount of labeled log data, unsupervised or self-supervised learning-based traditional methods are proposed. Lou et al. \cite{lou2010mining} introduced Invariant Mining (IM) to derive linear relations of the number of log templates, which can capture the co-occurrence patterns of log templates. Xu et al. \cite{xu2009detecting} applied Principal Component Analysis (PCA) to log anomaly detection, which uses the count vector of log templates and the parameter value vector of log messages as the inputs. Lin et al. \cite{lin2016log} used the Agglomerative Hierarchical Clustering algorithm (AHC) to classify log sequences into normal and anomalous clusters, which detects whether a newly arriving log sequence is anomalous by calculating its distance to the two clusters.

However, due to the differences in the distribution of statistical features of different datasets, these methods fail to achieve good performance on different datasets. In addition, when an anomaly occurs, a series of program executions are logged in the system, while traditional methods fail to capture the sequential dependencies in log sequences.

\subsection{Neural Network-Based Methods}

With the development of deep learning, neural network-based methods began to emerge. These methods usually capture the sequential dependencies in log sequences by infusing RNNs and their variants, e.g., LSTMs and GRUs. To portray the differences in patterns between normal and anomalous sequences, these methods model the patterns of log sequences into a unified feature representation. Du et al. \cite{du2017deeplog} predicted the log template of the next log message by employing LSTMs to learn the patterns of log sequences when the system runs normally. To capture the possible anomalies in the log parameters, the same LSTMs were applied to check the parameter values of log messages. Zhang et al. \cite{zhang2019robust} exploited bidirectional LSTMs to detect sequential anomalies with semantic features extracted by TextBoxes models \cite{liao2017textboxes}. Meng et al. \cite{meng2019loganomaly} captured quantitative anomalies by introducing the number of templates and detecting linear relations between log messages. Li et al. \cite{li2020swisslog} introduced time intervals in log sequences, and the bidirectional LSTMs were used to capture both the sequential dependencies and the patterns of time intervals.

To capture the complex log patterns, some methods enhance the perception of the features by introducing component sequences, auxiliary datasets, and multi-scale designs. Yin et al. \cite{yin2020improving} thought that the calling relations of components also contain log patterns and extracted the component sequences from log data as the inputs of the LSTMs. Nedelkoski et al. \cite{nedelkoski2020self} introduced auxiliary datasets from other systems as the negative sample set, and assumed that the effectiveness of log anomaly detection methods mainly depended on the ability of the model to distinguish between normal and anomalous sequences. Wang et al. \cite{wang2021multi} introduced a multi-scale design to slice the log sequence into fixed-length subsequences containing local log patterns, and LSTMs were used to extract the sequential dependencies at different scales.

Besides RNNs, other deep neural networks, e.g., GANs, CNNs, and GNNs, are also employed to address different challenges. Han et al. \cite{han2021unsupervised} and Xia et al. \cite{xia2021loggan} introduced GANs to generate new log data, which can enhance the ability of the model to capture anomalous information. To model the temporal correlation of log templates, Zhang et al. \cite{zhang2021logattn} and Wang et al. \cite{wang2022lightlog} arranged log templates as a matrix and captured the temporal dependencies in log sequences with CNNs and their variants. Wan et al. \cite{wan2021glad} transferred the log sequence into a graph, where nodes represent log templates and edges represent the order of log templates in the sequence, and GNNs were employed to model the information interactions in the graph and capture the log patterns. However, these methods ignore the interactions of subsequences, which can demonstrate the complex log patterns of components.

In this work, we propose CSCLog, a Component Subsequence Correlation-Aware Log anomaly detection method, which extracts subsequences from log sequences based on components and introduces an implicit correlation encoder to model their implicit correlations adaptively.

\section{Definitions and Preliminaries}
In this section, we provide the definitions of the associated terms used in CSCLog and formulate the template prediction task and anomaly detection task.

\textbf{Definition 1. Log Message.} A log message can be formalized as a tuple of attributes, denoted as $m=(e,p,t)$, where $e$ denotes the log template, which indicates the parsed structured textual data of the log data, $p$ denotes the component, and $t$ denotes the timestamp.

\textbf{Definition 2. Log Sequence.} A log sequence is regarded as a series of log messages ordered in time, denoted as $S = \left\{ m_{1},m_{2},\ldots,m_{N} \right\}$, where $N$ denotes the length of the log sequence.

\textbf{Definition 3. Subsequence.} A subsequence is regarded as a series of log messages that contain the same component, denoted as $P_{j} = \left\{ m_{1},m_{2},\ldots,m_{N_{p_{j}}} \right\}$, where $p_{j}$ and $N_{p_{j}}$ denote the component and the length of the subsequence, respectively.

\textbf{Problem 1. Template Prediction Task.} The template prediction task aims to predict the log template of the next log message, formalized as $\hat{e} = f\left( {S,\varepsilon_{\rm e},\varepsilon_{\rm p}} \right)$, where $S$, $\varepsilon_{\rm e}$, and $\varepsilon_{\rm p}$ denote the input log sequence, the set of log templates, and the set of components, respectively.

\textbf{Problem 2. Anomaly Detection Task.} Given a set of log sequences $M = \left\{ S_{1},S_{2},\ldots,S_{N_{\rm s}} \right\}$, where $N_{\rm s}$ denotes the number of the log sequences and each log sequence $S_{i}$ is normal. The anomaly detection task aims to detect whether a new log sequence $\hat{S}$ is normal or not by capturing the log patterns in $M$.

\section{Methodology}
In this section, we first give the framework of CSCLog and then describe individual modules in detail.

\begin{figure*}[!t]
\centerline{\includegraphics[width = \columnwidth]{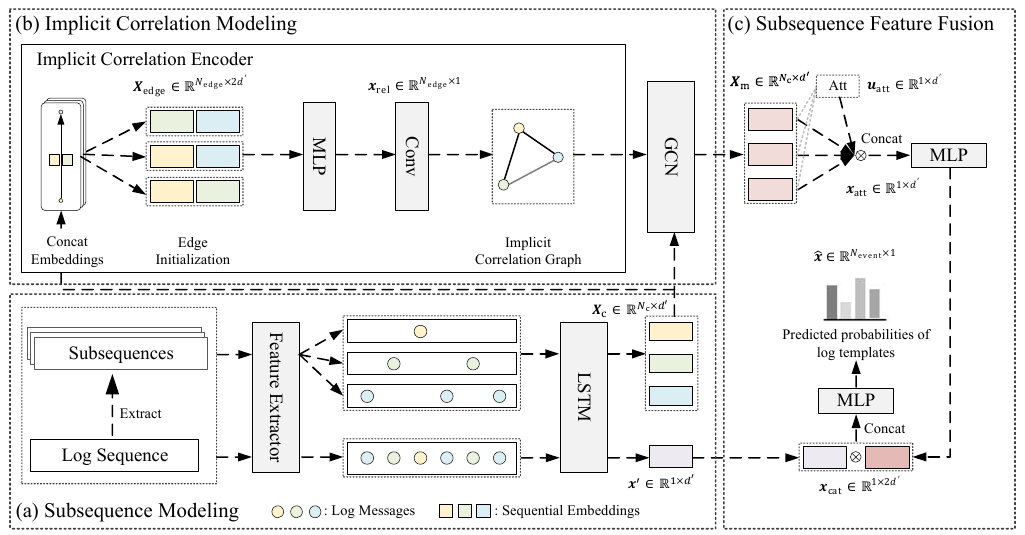}}
\caption{The overall framework of CSCLog.}
\label{fig2}
\end{figure*}

\subsection{Overview}
The framework of CSCLog is shown in Fig. \ref{fig2}, which consists of three modules: (1) subsequence modeling; (2) implicit correlation modeling; (3) subsequence feature fusion. For the subsequence modeling module, subsequences are extracted from the log sequence based on components, and LSTMs are introduced to capture the sequential dependencies in subsequences. For the implicit correlation modeling module, an implicit correlation encoder is introduced to model the implicit correlations of subsequences adaptively, and GCNs are employed to accomplish the interactions of subsequences, which can model the influences between subsequences. For the subsequence feature fusion module, attention mechanisms are exploited to fuse the embeddings of all subsequences. The details of these modules are introduced in Sections \ref{sm2} to \ref{sm4}.

\subsection{Subsequence Modeling}\label{sm2}
We extract subsequences from the log sequence, where the log messages in the same subsequence contain the same component, and create empty subsequences for components that are not contained in the log sequence. We denote the set of subsequences as $C = \left\{ P_{1},P_{2},\ldots,P_{N_{\rm c}} \right\}$, where $N_{\rm c}$ is the number of components.

To capture anomalous information with different features in log data, we extract semantic and temporal features in log sequences and subsequences. The semantic features are the descriptions of the unstructured text in the log templates. We first create a set of keywords extracted from the text of the log templates by removing the non-character tokens and prepositions. Then, to obtain semantic information of the words in the set, we map each word to a 768-dimensional word vector by a pre-trained BERT model \cite{sun2019bert4rec}. Finally, we calculate the weighted average of the vectors of all keywords in the log template to obtain its semantic vector $\boldsymbol{v}^\prime$, which is formalized as:
\begin{equation}
\boldsymbol{v}^\prime=\frac{1}{N_{\rm e}}\sum_{i=1}^{N_{\rm e}}w_i\boldsymbol{v}_i\label{eq1}
\end{equation}
where $\boldsymbol{v}_i$, $w_i$, and $N_{\rm e}$ denote the word vector of the $i${th} keyword, the Term Frequency-Inverse Document Frequency (TF-IDF) weight of the $i${th} keyword, and the number of keywords in the log template, respectively. By extracting semantic features for all log messages in the log sequence, we obtain its semantic features $\boldsymbol{V}\in\mathbb{R}^{N\times768}$ of the log sequence, where $N$ denotes the length of the sequence.

Temporal features focus on extracting the difference in time of log messages. Specifically, the timestamp of the first log message in the log sequence is regarded as the start time, and we take the interval of the timestamp of a log message and the start time as its temporal feature $t^\prime$, which is formalized as:
\begin{equation}
t^\prime=t_i-t_1\label{eq2}
\end{equation}
where $t_i$ denotes the timestamp of the $i${th} log message, $t_i\in\mathbb{N}$. By extracting the temporal features for all log messages in the log sequence, we obtain its temporal features $\boldsymbol{t}\in\mathbb{R}^{N\times1}$.

We introduce a feature extractor to map features with different dimensions into a low-dimensional feature space. Semantic and temporal features are first sent to a corresponding Multilayer Perceptron (MLP) to obtain semantic and temporal embeddings, whose dimensions are $N\times d_{\rm sem}$ and $N\times d_{\rm tim}$, respectively. Then, we concatenate these two embeddings to obtain the feature embedding $\boldsymbol{X}\in\mathbb{R}^{N\times d}$, which is formalized as:
\begin{equation}
\boldsymbol{X}={\rm Concat}\left(\varphi_1\left(\boldsymbol{V}\right),\varphi_2\left(\boldsymbol{t}\right)\right)\label{eq3}
\end{equation}
where $\varphi_1$ and $\varphi_2$ denote embedding operations, which are accomplished by MLP, and ${\rm Concat}$ denotes the concatenation operation. In addition, to balance the effects of semantic and temporal features on model performance, a threshold $\alpha_{\rm emb}\in(0,1)$ is applied to control the dimensions of the two embeddings, i.e., $d_{\rm sem}=\alpha_{\rm emb}d$ and $d_{\rm tim}=(1-\alpha_{\rm emb})d$.

Finally, we employ the LSTMs to capture the sequential dependencies in the log sequence and obtain its sequential embedding $\boldsymbol{x}^\prime\in\mathbb{R}^{1\times d^\prime}$, which is formalized as:
\begin{equation}
\boldsymbol{x}^\prime={\rm LSTM}\left(\boldsymbol{X}\right)\label{eq4}
\end{equation}
Similarly, we obtain the sequential embeddings of all the subsequences $\boldsymbol{X}_{\rm c}\in\mathbb{R}^{N_{\rm c}\times d^\prime}$, where $N_{\rm c}$ denotes the number of components. The parameters of the LSTMs are shared between subsequences.

\subsection{Implicit Correlation Modeling}\label{sm3}
To model the interactions of subsequences, we introduce the implicit correlation encoder to learn the implicit correlation of different subsequences end-to-end. Specifically, to avoid the correlation of any subsequence being lost, the embeddings of subsequences are concatenated in pairs based on the assumption that interactions exist between any subsequence pair \cite{kipf2018neural}. We obtain the correlation embedding $\boldsymbol{X}_{\rm edge}\in\mathbb{R}^{N_{\rm edge}\times2d^\prime}$, where $N_{\rm edge}$ denotes the number of subsequence pairs. The correlation embedding of a subsequence pair is formalized as:
\begin{equation}
\boldsymbol{X}_{\rm edge}^{i,j}=\sigma_1\left(\varphi_3\left({\rm Concat}\left(\boldsymbol{X}_{\rm c}^{i},\boldsymbol{X}_{\rm c}^{j}\right)\right)\right)\label{eq5}
\end{equation}
where $\boldsymbol{X}_{\rm c}^{i}$ and $\boldsymbol{X}_{\rm c}^{j}$ denote the embedding of the $i${th} and $j${th} subsequences, respectively, $\sigma_1$ denotes the ReLU activation function, and $\varphi_3$ denotes the MLP. Then, we obtain the correlation weight $\boldsymbol{x}_{\rm rel}\in\mathbb{R}^{N_{\rm edge}\times1}$ of each subsequence pair based on their correlation embeddings, which is formalized as:
\begin{equation}
\boldsymbol{x}_{\rm rel}^{i,j}=\sigma_2\left({\rm Conv}\left(\boldsymbol{X}_{\rm edge}^{i,j}\right)\right)\label{eq6}
\end{equation}
where $\boldsymbol{X}_{\rm edge}^{i,j}$ denotes the correlation embedding of the $i${th} and $j${th} subsequences, ${\rm Conv}$ denotes the convolution operation, and $\sigma_2$ denotes the Sigmoid activation function, which remains the correlation weight between 0 and 1.

Finally, GCNs are employed to accomplish the interactions of subsequences. Specifically, we construct the implicit correlation graph, in which the embeddings of subsequences are regarded as the embeddings of nodes in the graph, and the correlation weights between different subsequences are regarded as the weights of edges, which indicate the importance of the interactions of subsequences. The updated subsequence embedding $\boldsymbol{X}_{\rm m}\in\mathbb{R}^{N_{\rm c}\times d^\prime}$ is obtained by stacking the GCN layers, which is formalized as:
\begin{equation}
\boldsymbol{X}_{\rm m}=\sigma_3\left({\rm GCN}\left(\boldsymbol{X}_{\rm c},\boldsymbol{x}_{\rm rel}\right)\right)\label{eq7}
\end{equation}
where $\boldsymbol{X}_{\rm c}$ denotes embeddings of subsequences, $\boldsymbol{x}_{\rm rel}$ denotes the correlation weight, ${\rm GCN}$ denotes the message passing operation, and $\sigma_3$ denotes the ReLU activation function.

\subsection{Subsequence Feature Fusion}\label{sm4}
Attention mechanisms are employed to fuse the embeddings of different subsequences without considering the effect of positional relations. Specifically, we introduce a global vector $\boldsymbol{u}_{\rm att}\in\mathbb{R}^{1\times d^\prime}$, and the attention weights $\beta_i$ of the embeddings of different subsequences are calculated by the attention mechanisms \cite{chen2021multiscale}, which is formalized as:
\begin{equation}
\beta_i=\frac{{\rm AttScore}\left(\boldsymbol{X}_{\rm m}^{i},\boldsymbol{u}_{\rm att}\right)}{\sum_{k=0}^{N_{\rm c}}\left({\rm AttScore}\left(\boldsymbol{X}_{\rm m}^{k},\boldsymbol{u}_{\rm att}\right)\right)}\label{eq8}
\end{equation}
where ${\rm AttScore}$ denotes the attention scoring function, which is accomplished by the dot product of vectors, and $\boldsymbol{X}_{\rm m}^{i}$ denotes the embedding of the $i${th}  subsequence. To avoid the weight $\beta_i$ jumping between two neighboring iterations, resulting in unstable training, we introduce a constraint on it, which is formalized as:
\begin{equation}
\beta_i^j\gets\gamma\beta_i^{j-1}+(1-\gamma)\beta_i^j\label{eq9}
\end{equation}
where $\gamma$ denotes the dynamic threshold and $\beta_i^j$ denotes the attention weight of the $i${th} subsequence at the $j${th} iteration. Then, we fuse the embeddings of all subsequences to get the fused embedding $\boldsymbol{x}_{\rm att}\in\mathbb{R}^{1\times d^\prime}$, which is formalized as:
\begin{equation}
\boldsymbol{x}_{\rm att}=\sum_{i=0}^{N_{\rm c}}{\beta_i\boldsymbol{X}_{\rm m}^{i}}\label{eq10}
\end{equation}
where $\boldsymbol{X}_{\rm m}^{i}$ denotes the embedding of the $i${th} subsequence and $\beta_i$ denotes the attention weight of the $i${th} subsequence.

Finally, to obtain the probabilities of the log template of the next log message $\hat{\boldsymbol{x}}\in\mathbb{R}^{N_{\rm event}\times1}$, where $N_{\rm event}$ denotes the number of log templates, we concatenate the fused embedding with the sequential embedding of the log sequence and send it to the classifier, which is formalized as:
\begin{equation}
\hat{\boldsymbol{x}}=\sigma_{\rm out}\left(\varphi_{\rm out}\left({\rm Concat}\left(\boldsymbol{x}^\prime,\boldsymbol{x}_{\rm att}\right)\right)\right)\label{eq11}
\end{equation}
where $\boldsymbol{x}^\prime$ denotes the sequential embedding, $\boldsymbol{x}_{\rm att}$ denotes the fused embedding of subsequences, $\varphi_{\rm out}$ denotes the transformation function, which is implemented by the MLP, and $\sigma_{\rm out}$ denotes the Softmax activation function.

\subsection{Training and Anomaly Detection}
CSCLog is trained on the template prediction task. The training objective is to minimize the cross-entropy loss, which is formalized as:
\begin{equation}
\mathcal{L}=-\sum_{i=1}^{N_{\rm event}}{y_i\log({\hat{\boldsymbol{x}}}_{i})}\label{eq12}
\end{equation}
where ${\hat{\boldsymbol{x}}}_{i}$ denotes the predicted probability of the $i$th log template, and $y_i$ denotes whether the predicted log template is the same as the true log template, which is formalized as:
\begin{equation}
y_{i} = \left\{ \begin{matrix}
{1,~~i = {\rm index~of~the~true~log~template}} \\
{0,~~{\rm otherwise}}\quad\quad\quad\quad\quad\quad\quad\quad\quad\,\,\,\\
\end{matrix} \right.\label{eq13}
\end{equation}

To improve the efficiency of the calculation of the model, the input data are calculated in parallel using the batch processing method.

To determine whether a log sequence is normal or anomalous, we use a sliding window to get a set of log sequence samples. Each sample $S_{\rm t} = \left\{ m_{1},m_{2},\ldots,m_{N_{\rm w}}\right\}$, where $N_{\rm w}$ denotes the length of the sliding window, is sent to the trained model to get the set of top-\textit{k} ranked  templates. $S_{\rm t}$ is regarded as an anomaly sample when the true log template of the next log message is not in the set, which is formalized as:
\begin{equation}
y_{\rm t} = \left\{ \begin{matrix}
{0,~~e_{\rm t} \in {\rm the~set~of~top\text{-}\textit{k}~ranked~templates}} \\
{1,~~{\rm otherwise}}\quad\quad\quad\quad\quad\quad\quad\quad\quad\quad\quad\,\,\\
\end{matrix} \right.\label{eq14}
\end{equation}
where $e_{\rm t}$ denotes the true log template of the next log message. We consider that an anomaly occurs in the log sequence when the number of anomalous samples reaches the set threshold $\alpha_{\rm anom}$.


\begin{table*}[!t]
    \caption{The details of the experimental datasets.}
    \begin{center}
    \setlength{\tabcolsep}{3mm}{
    \resizebox{\linewidth}{!}{
    \begin{tabular}{cccccc}
        \toprule
        Datasets    &   \# Components    &  \# Log templates &  \# Log sequences &  Average length &    Anomaly rate    \\
        \midrule
        HDFS	&  8	& 47	&    22,023	&    32.9	&  9.1\% \\ 
        BGL	&   10	&    1,372	& 13,074	&    62.3	&  30.6\%    \\
        ThunderBird	&   48	&    302	&   12,841	&    37.8	&  22.4\%  \\
        OpenStack	& 12	&    42	&    5,753	& 28.6	&  9.6\%   \\
        \bottomrule
    \end{tabular}}
}
\label{tab1}
\end{center}
\vspace{5pt}
\end{table*}

\begin{table*}[!t]
    \caption{Search spaces of hyper-parameters and configures of NNI.}
    \begin{center}
    \setlength{\tabcolsep}{3mm}{
    \resizebox{\linewidth}{!}{
    \begin{tabular}{cccc}
        \toprule
        	&  Parameters	&    Datasets	&  Choice   \\
        \midrule
        \multirow{11}{*}{Search spaces}   &   Learning rate	& All	&   \{0.0001, 0.001, 0.01\} \\
            &   Batch size	&    All	&   \{8, 16, 32\} \\
            &   Threshold of feature extractor	&    All	&   \{0.2, 0.5, 0.8\}   \\
            &   \# Hidden layer units    &	All	& \{128, 256, 512\}   \\
            &   Dropout rate	&  All	&   \{0.1, 0.5, 0.9\}   \\
            &   Length of sliding window	&  All	&   \{9, 11, 13, 15, 17, 19, 21 23, 25\}    \\
            &   Threshold of anomaly detection	&    All	&   \{1, 2, 3, 4, 5, 6, 7, 8, 9\}   \\
            &   \multirow{4}{*}{\# Top-\textit{k} ranked templates}	&  HDFS	&  \{1, 3, 5, 7, 9, 11, 13, 15, 17\}   \\
            &&  BGL	&   \{1, 23, 45, 67, 89, 111, 133, 155, 177\}   \\
            &&  ThunderBird	&   \{1, 5, 9, 13, 17, 21, 25, 29, 33\} \\
            &&  OpenStack	& \{1, 3, 5, 7, 9, 11, 13, 15, 17\}   \\
        \cmidrule{2-4}
       \multirow{3}{*}{Configures of NNI}	& Max trial number	&  All	&   100 \\
            &   Optimization algorithm	&    All	&   TPE \\
            &   Parallel number	&   All	&   2   \\
        \bottomrule
    \end{tabular}}
    }
\label{tab2}
\end{center}
\vspace{5pt}
\end{table*}

\begin{table*}[!t]
    \caption{The best hyper-parameters on different datasets.}
    \begin{center}
    \setlength{\tabcolsep}{6mm}{
    \begin{tabular}{ccccc}
        \toprule
        Parameters  &  HDFS	& BGL	&    ThunderBird	&    OpenStack   \\
        \midrule
        Learning rate   &   0.0001  &   0.0001  &   0.0001  &   0.0001  \\
        Batch size      &   16      &   8       &   32      &   32      \\
        Threshold of feature extractor  &   0.8 &   0.5 &   0.8 &   0.8 \\
        \# Hidden layer units   &   512 &   256 &   128 &   256     \\
        Dropout rate    &   0.1 &   0.9 &   0.5 &   0.1 \\
        Length of sliding window    &   19  &   25  &   11  &   9   \\
        Threshold of anomaly detection  &   4   &   8   &   1   &   4   \\
        \# Top-\textit{k} ranked templates   &   7   &   45  &   29  &   3   \\
        \bottomrule
    \end{tabular}}
\label{tab3}
\end{center}
\vspace{5pt}
\end{table*}

\section{Experiments}
In this section, we justify the superiority of CSCLog by conducting extensive experiments. Firstly, we introduce the experimental datasets and settings. Subsequently, we present the comparison with baselines, the ablation study, the parameter sensitivity analysis, and the case study.

\subsection{Datasets}
To evaluate the performance of CSCLog, we conduct experiments on four public log datasets, i.e., HDFS \cite{xu2009detecting}, BGL \cite{oliner2007supercomputers}, ThunderBird \cite{oliner2007supercomputers}, and OpenStack \cite{du2017deeplog}, which contain anomaly labels. The details of the experimental datasets are shown in Table \ref{tab1}.

\textbf{HDFS}: The HDFS dataset is generated by Hadoop-based MapReduce jobs deployed on more than 2,000 Amazon’s EC2 nodes, which contains 11,175,629 log messages for 39 hours. The log sequences are extracted directly based on the block\_id in a log message, which are manually labeled as anomaly or normal by the Hadoop domain experts.

\textbf{BGL}: The BGL dataset contains 4,747,963 log messages generated by the BlueGene/L supercomputer deployed at Lawrence Livermore National Laboratory, with a time span of 7 months. Each log message in the dataset is manually labeled as anomaly or normal by the domain experts. A sliding window with the length of 10 seconds is used to extract the log sequences. Log sequences are labeled as anomaly when containing anomalous log messages.

\textbf{ThunderBird}: ThunderBird is a large dataset of over 200 million log messages, generated on the ThunderBird supercomputer system at Sandia National Laboratories (SNL). We extracted 2 million continuous log messages from 9:18:40 on November 11, 2005 to 20:28:59 on November 13, 2005, with a time span of about 59 hours. Similar to BGL, each log message is manually labeled as anomaly or normal by the domain experts. A sliding window with the length of 10 seconds is used to extract the log sequences. Log sequences are labeled as anomaly when containing anomalous log messages.

\textbf{OpenStack}: OpenStack is a small dataset generated by a cloud operating system with 10 compute nodes, which contains 207,636 log messages for 30 hours. The dataset is divided into two normal log files and an anomaly file. A sliding window with the length of 10 seconds is used to extract the log sequences, and the log sequence is labeled with the same label as the file.

\subsection{Experimental Settings}
CSCLog is implemented in Python with PyTorch 1.10.0 \cite{paszke2019pytorch} and PyTorch Geometric 2.1.0 \cite{rozemberczki2021pytorch}, and the source code is released on GitHub\footnote{https://github.com/Hang-Z/CSCLog}. We conduct all the experiments with an Intel Xeon Gold 5118 CPU and an NVIDIA GeForce RTX 2080Ti GPU. We parse the log messages by a parsing tool Drain \cite{he2017drain} with default parameters. All the datasets are split into training, validation, and test sets by the ratio of 7:1:2 according to the temporal order of log sequences.

Adam \cite{kingma2015adam} is adopted as the optimizer, and the weight decay rate is set to 0.0001. The number of iterations is set to 20. To avoid overfitting, we apply the early stop \cite{yao2007early} in the training stage, where the stop patience is set to 50\% of the number of iterations by default. The number of layers of the GNNs in implicit correlation encoder is set to 2. The number of layers of the LSTMs in subsequence modeling module is set to 2. The step length of the sliding window for sampling is set to 1. The global vector $\boldsymbol{u}_{\rm att}$ in subsequence feature fusion module is initialized with the Xavier initialization. For other hyper-parameters in the model, we exploit the Neural Network Intelligence toolkit (NNI)\footnote{https://nni.readthedocs.io/en/latest/}  to search for the best value automatically. Due to the differences in the number of log templates for different datasets, the hyper-parameter number of top-\textit{k} ranked templates is chosen differently for them. The search spaces of these hyper-parameters and the configures of NNI are given in Table \ref{tab2}. The best hyper-parameters on different datasets are given in Table \ref{tab3}.

We use Accuracy, Precision, Recall, and Macro F1-Measure as evaluation metrics. Macro F1-Measure combines Precision and Recall and allocates same weight to every class, which reduces the impact of class imbalance on the evaluation results, and higher Macro F1-Measure demonstrates better performance, which is formalized as:
\begin{equation}
{\rm F1}=\frac{1}{N_{\rm class}}\times\sum_{k=1}^{N_{\rm class}}\frac{2\times{\rm Precision}_{k}\times{\rm Recall}_{k}}{{\rm Precision}_{k}+{\rm Recall}_{k}}\label{eq15}
\end{equation}
where $k$ denotes the index of classes and $N_{\rm class}$ denotes the number of classes. The final reported result is the average macro F1-Measure across five independent runs by varying seeds.

\begin{table*}[!t]
    \caption{Comparisons with baselines on the anomaly detection task. The best results are bold and the second best results are underlined.}
    \begin{center}
    \setlength{\tabcolsep}{0.1mm}{
    \resizebox{\linewidth}{!}{
    \begin{tabular}{ccccccccccccccccc}
        \toprule
        \multirow{2.5}{*}{Model}  &  \multicolumn{4}{c}{HDFS}	& \multicolumn{4}{c}{BGL}	&    \multicolumn{4}{c}{ThunderBird}	&    \multicolumn{4}{c}{OpenStack}   \\
        \cmidrule{2-17}
        &   \textit{Acc}.	&  \textit{Pre}.	&  \textit{Rec}.	&  \textit{F1}.	&   \textit{Acc}.	&  \textit{Pre}.	&  \textit{Rec}.	&  \textit{F1}. &   \textit{Acc}.	&  \textit{Pre}.	&  \textit{Rec}.	&  \textit{F1}. &   \textit{Acc}.	&  \textit{Pre}.	&  \textit{Rec}.	&  \textit{F1}. \\
        \midrule
        IM (USENIX ATC 2010)	&    0.897	& \underline{0.735}	& 0.944	& 0.790  &0.712	&0.729	&0.538	&0.492 &0.889	&0.832	&0.877	&0.851 &0.866	&0.482	&0.491	&0.481    \\
        PCA (SOSP 2009)	&0.900	&0.738	&\underline{0.945}	&\underline{0.793} &0.660	&0.415	&0.482	&0.416	&0.960	&0.929	&0.966	&0.945 &0.892	&0.485	&0.498	&0.479     \\
        LogCluster (ICSE 2016)	&\underline{0.906}	&0.454	&0.498	&0.475 & 0.551	&0.617	&0.626	&0.550	&0.776	&0.388	&0.500	&0.437 &\underline{0.904}	&0.452	&0.500	&0.475 \\
        DeepLog (CCS 2017)	&0.895	&0.732	&0.942	&0.786 &0.481	&0.670	&0.621	&0.473	&0.962	&0.938	&0.956	&0.947 &0.897	&0.507	&\underline{0.501}	&0.481 \\
        LogAnomaly (IJCAI 2019)	&0.895	&0.734	&0.942	&0.787  &0.479	&0.670	&0.619	&0.471	&0.964	&0.941	&0.958	&0.949 &0.898	&\underline{0.514}	&\underline{0.501}	&\underline{0.482} \\
        LogC (ICSME 2020)	&0.898	&0.735	&0.944	&0.790   & \underline{0.727}	&\textbf{0.858}	&0.593	&0.582	&0.504	&0.643	&0.671	&0.502 &0.874	&0.477	&0.492	&0.480 \\
        OC4Seq (KDD 2021)	&0.899	&0.736	&0.944	&0.791  &0.707	&0.696	&\underline{0.730}	&\underline{0.692}	&\underline{0.974}	&\underline{0.948}	&\textbf{0.972}	&\underline{0.956} &0.903	&0.452	&0.500	&0.474 \\
        CSCLog (Ours)	&\textbf{0.966}	&\textbf{0.824}	&\textbf{0.952}	&\textbf{0.895}  &\textbf{0.739}	&\underline{0.749}	&\textbf{0.793}	&\textbf{0.731}	&\textbf{0.982}	&\textbf{0.978}	&\underline{0.969}	&\textbf{0.97}3 &\textbf{0.906}	&\textbf{0.535}	&\textbf{0.528}	&\textbf{0.530}   \\
        \bottomrule
    \end{tabular}}
    }
\label{tab4}
\end{center}
\vspace{2pt}
\end{table*}

\begin{table*}[!t]
    \caption{Comparisons with baselines on template prediction task. The best results are bold.}
    \begin{center}
    \setlength{\tabcolsep}{1.2mm}{
    \resizebox{\linewidth}{!}{
    \begin{tabular}{ccccccccccccc}
        \toprule
        \multirow{2.5}{*}{Model}  &  \multicolumn{4}{c}{$\textit{k} = 1$}	& \multicolumn{4}{c}{\textit{k} = 3}	&    \multicolumn{4}{c}{\textit{k} = 5}   \\
        \cmidrule{2-13}
        &   \textit{Acc}.	&  \textit{Pre}.	&  \textit{Rec}.	&  \textit{F1}.	&   \textit{Acc}.	&  \textit{Pre}.	&  \textit{Rec}.	&  \textit{F1}. &   \textit{Acc}.	&  \textit{Pre}.	&  \textit{Rec}.	&  \textit{F1}.  \\
        \midrule
        DeepLog (CCS 2017)	&0.904	&0.215	&0.187	&0.193	&0.983	&0.325	&0.305	&0.312	&0.987	&0.327	&0.333	&0.330   \\
        LogAnomaly (IJCAI 2019)	&0.909	&0.220	&0.196	&0.199	&0.983	&0.261	&0.273	&0.267	&0.987	&0.348	&0.344	&0.341  \\
        LogC (ICSME 2020)	&0.902	&0.213	&0.187	&0.192	&0.983	&0.326	&0.308	&0.314	&0.987	&0.327	&0.333	&0.330    \\
        OC4Seq (KDD 2021)	&0.910	&0.226	&0.201	&0.203	&0.981	&0.245	&0.268	&0.256	&0.987	&0.335	&0.352	&0.342  \\
        CSCLog (Ours)	&\textbf{0.911}	&\textbf{0.230}	&\textbf{0.210}	&\textbf{0.208}	&\textbf{0.988}	&\textbf{0.359}	&\textbf{0.364}	&\textbf{0.361}	&\textbf{0.989}	&\textbf{0.360}	&\textbf{0.364}	&\textbf{0.362}  \\
        \bottomrule
    \end{tabular}}
    }
\label{tab5}
\end{center}
\end{table*}

\subsection{Comparison with Baselines}
To justify the superiority of CSCLog, we compare it with other log anomaly detection methods, all of which are unsupervised or self-supervised learning-based methods. Among these methods, Invariant Mining (IM) \cite{lou2010mining}, Principal Components Analysis (PCA) \cite{xu2009detecting}, and LogCluster \cite{lin2016log} are traditional methods, and DeepLog \cite{du2017deeplog}, LogAnomaly \cite{meng2019loganomaly}, LogC \cite{yin2020improving}, and OC4Seq \cite{wang2021multi} are neural network-based methods. Due to the additional introduction of anomalous information, supervised learning-based methods usually perform better than unsupervised or self-supervised learning-based methods \cite{le2022log}, so that they are not considered in our comparison experiment. We conduct experiments on the anomaly detection task and template prediction task, respectively. All compared methods are implemented by the code provided by the original papers. For the sake of fairness, all compared methods follow the same experimental settings and search spaces of hyper-parameters as CSCLog, and their parameters are also optimized. In addition, the kernel size of the one-class classifier of OC4Seq is set to 5 according to the original paper. The detailed descriptions of baselines are as follows:

\textbf{IM} \cite{lou2010mining}, \textbf{PCA} \cite{xu2009detecting}, and \textbf{LogCluster} \cite{lin2016log}: IM, PCA, and LogCluster are traditional methods that detect anomalies by extracting the statistical features in log sequences. To use them on the log anomaly detection task, we extract the number of log templates for IM and PCA, and construct the set of frequent words in log messages for LogCluster. To a newly arriving log sequence, we detect whether it is anomalous according to the patterns captured from the statistical features.

\textbf{DeepLog} \cite{du2017deeplog}: DeepLog predicts the log template of the next log message by modeling the log sequences as natural language sequences and employing LSTMs to learn the patterns of log sequences when the system runs normally. To a  newly arriving log sequence, DeepLog detects whether it is anomalous according to whether the true log template is in the predicted log templates.

\textbf{LogAnomaly} \cite{meng2019loganomaly}: LogAnomaly captures quantitative anomalies by introducing the count vector of log templates in log sequences, and captures sequential anomalies by learning the patterns of log sequences when the system runs normally. Similar  to DeepLog, LogAnomaly detects whether a newly arriving log sequence is anomalous according to whether the true log template is in the predicted log templates.

\textbf{LogC} \cite{yin2020improving}: LogC extracts the component sequences from log data, and captures the calling relations of components and the patterns of log sequences when the system runs normally. To  a newly arriving log sequence, LogC detects whether it is anomalous according to whether the true log template is in the predicted log templates and the true component is in the predicted components.

\textbf{OC4Seq} \cite{wang2021multi}: OC4Seq introduces a multi-scale design to slice the log sequence into fixed-length subsequences containing local log patterns. LSTMs are used to capture the sequential dependencies at different scales. A one-class  classifier is introduced to detect whether a newly arriving log sequence is anomalous by capturing the features of sequential embeddings.

Table \ref{tab4} shows the Accuracy, Precision, Recall, and Macro F1-Measure of CSCLog and baselines on the anomaly detection task on HDFS, BGL, ThunderBird, and OpenStack datasets, from which we can observe the following phenomena:

1) Neural network-based methods (DeepLog, LogAnomaly, LogC, and OC4Seq) outperform traditional methods (IM, PCA, and LogCluster), which indicates the effectiveness of capturing the sequential dependencies in log sequences. In addition, the high differences in the performances of traditional methods on different datasets demonstrate the limitations of extracted statistical features.

2) The performance of LogAnomaly is slightly better than that of DeepLog, which indicates that the introduced number of templates can enhance the ability to capture anomalous information. However, LogAnomaly fails to perform well on the BGL dataset. The reason is that the large number of log templates (over 1000) contributes to the sparse count vector of templates, so that the model fails to capture the linear relations between log templates.

3) LogC fails to perform well on the ThunderBird dataset, which contains a large number of components. The reason is that LogC focuses on modeling the calling relations of components, and the large number of components make the relations more complex, which presents a challenge to the model to capture the complex log patterns of components.

4) As the SOTA method on the anomaly detection task, OC4Seq performs better than DeepLog, LogAnomaly, and LogC. OC4Seq can capture both global and local sequential dependencies in log sequences by slicing the subsequences of different lengths. In addition, by introducing the one-class classifier to classify log sequences directly, OC4Seq reduces the impact of the length of sequences on performance.

5) CSCLog outperforms the best baseline with relative improvements of 12.86\%, 5.64\%, 1.19\%, and 9.96\% in Macro F1-Measure on HDFS, BGL, ThunderBird, and OpenStack datasets, respectively. The results justify the advantage of capturing the sequential dependencies in subsequences and modeling the implicit correlations of subsequences.

Table \ref{tab5} shows the Accuracy, Precision, Recall, and Macro F1-Measure of CSCLog and neural network-based methods on the template prediction task on the HDFS dataset. To demonstrate the comprehensive results, we set the number of top-\textit{k} ranked templates to 1, 3, and 5. We replace the one-class classifier of OC4Seq with the same classifier as CSCLog to accomplish the template prediction task. We can observe that CSCLog outperforms all the baselines and performs better than the best baseline with relative improvements of 3.37\%, 14.97\%, and 5.85\% in Macro F1-Measure. This further demonstrates the advancement of CSCLog.

\begin{table}[!t]
\vspace{5pt}
    \caption{Comparisons with simplified models with different modules. The best results are bold.}
    \begin{center}
    \setlength{\tabcolsep}{2.8mm}{
    \begin{tabular}{ccccc}
        \toprule
        Model  &     \textit{Acc}.	&  \textit{Pre}.	&  \textit{Rec}.	&  \textit{F1}.  \\
        \midrule
        CSCLog w/o IC	&0.799	&0.666	&0.512	&0.495   \\
        CSCLog w/o LSTM	&0.838	&0.719	&0.640	&0.587    \\
        CSCLog	&\textbf{0.966}	&\textbf{0.824}	&\textbf{0.952}	&\textbf{0.895}  \\
        \bottomrule
    \end{tabular}}
\label{tab6}
\end{center}
\end{table}

\begin{table}[!t]
    \caption{Comparisons with simplified models with different features. The best results are bold.}
    \begin{center}
    \setlength{\tabcolsep}{2.8mm}{
    \begin{tabular}{ccccccccccccc}
        \toprule
        Model  &     \textit{Acc}.	&  \textit{Pre}.	&  \textit{Rec}.	&  \textit{F1}.  \\
        \midrule
        CSCLog w/o SEM	&0.825	&0.701	&0.653	&0.602   \\
       CSCLog w/o TIME	&0.877	&0.722	&0.668	&0.639    \\
        CSCLog	&\textbf{0.966}	&\textbf{0.824}	&\textbf{0.952}	&\textbf{0.895}  \\
        \bottomrule
    \end{tabular}}
\label{tab7}
\end{center}
\end{table}

\subsection{Ablation Studies}
To justify the advantage of capturing the sequential dependencies in subsequences and modeling the implicit correlations of subsequences, we compare CSCLog with two simplified models. For the sake of fairness, all variants follow the same experimental settings as CSCLog, and their parameters are also optimized. The detailed descriptions of simplified models are as follows:

\textbf{CSCLog w/o IC}: CSCLog w/o IC removes the implicit correlation encoder and sets the correlation weight between subsequences to 1, which means that the correlations of subsequences are fixed and identical. The rest is the same as CSCLog.

\textbf{CSCLog w/o LSTM}: CSCLog w/o LSTM removes the LSTMs of the subsequence modeling module from CSCLog and only utilizes the average of the feature embeddings of a log sequence as its sequential embedding, which means that the model cannot capture the sequential dependencies. The rest is the same as CSCLog.

Table \ref{tab6} shows the Accuracy, Precision, Recall, and Macro F1-Measure of CSCLog and simplified models on the anomaly detection task on the HDFS dataset, from which we can observe that CSCLog outperforms CSCLog w/o IC, which indicates the effectiveness of using the implicit correlation encoder to model the implicit correlation of subsequences. CSCLog outperforms CSCLog w/o LSTM, which indicates the effectiveness of employing LSTMs to capture the sequential dependencies in log sequences. The performance degradation of CSCLog w/o IC is more than that of CSCLog w/o LSTM, which indicates that modeling the implicit correlation of subsequences is more effective to improve the ability to capture anomalous information.

To justify the effectiveness of introduced features, we remove semantic and temporal features to study the impact on the model. CSCLog w/o SEM denotes the model without semantic features and CSCLog w/o TIME denotes the model without temporal features.

Table \ref{tab7} shows the Accuracy, Precision, Recall, and Macro F1-Measure of CSCLog and simplified models on the anomaly detection task on the HDFS dataset, from which we can observe that CSCLog outperforms CSCLog w/o SEM and CSCLog w/o TIME, which indicates the effectiveness of the introduced semantic and temporal features. The performance degradation of CSCLog w/o SEM is more than that of CSCLog w/o TIME, which indicates the importance of semantic features. The reason is that the semantic features contain rich text information of log templates.

\subsection{Parameter Sensitivity Analysis}
To study the impact of several important parameters, including the length of sliding window, the number of top-\textit{k} ranked templates, and the threshold of anomaly detection, we conduct the parameter sensitivity analysis on the anomaly detection task.

\begin{figure*}[ht]
    \centering
	\begin{minipage}[t]{0.32\linewidth}
    	\centering	
            \includegraphics[width=1\linewidth]{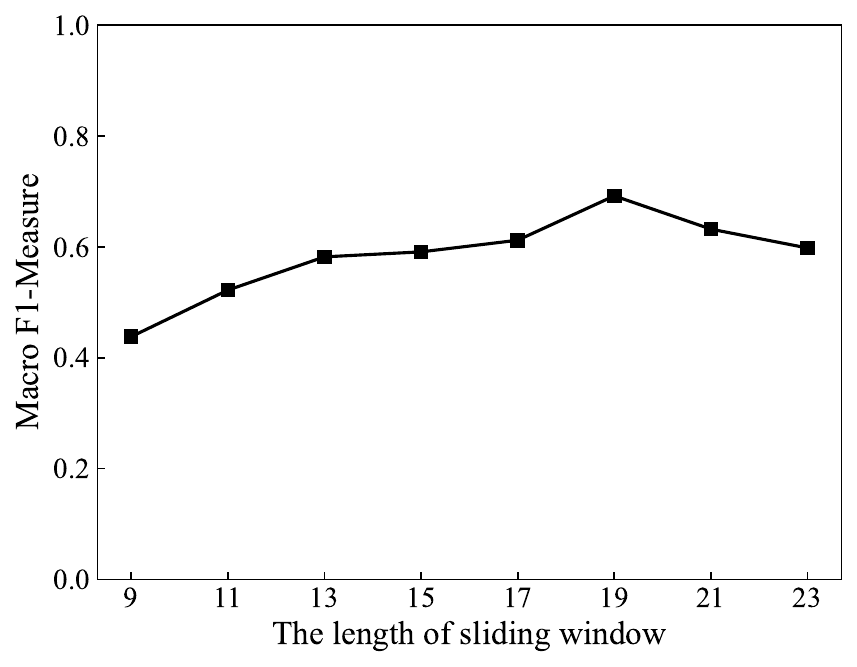}
            \vspace{-10pt}
            \caption{Impact of the length of sliding window.}
            \label{fig3}
	\end{minipage}%
        \hspace{0.01\linewidth}
	\begin{minipage}[t]{0.32\linewidth}
		\centering
            \includegraphics[width=1\linewidth]{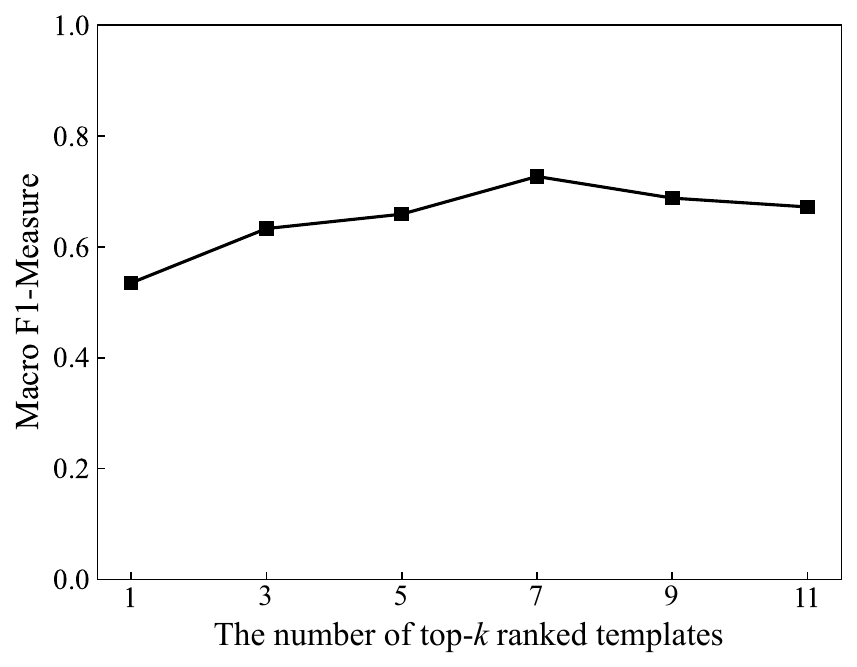}
            \vspace{-10pt}
            \caption{Impact of the number of top-$k$ ranked templates.}
            \label{fig4}
	\end{minipage}%
        \hspace{0.01\linewidth}
	\begin{minipage}[t]{0.32\linewidth}
		\centering
    		\includegraphics[width=1\linewidth]{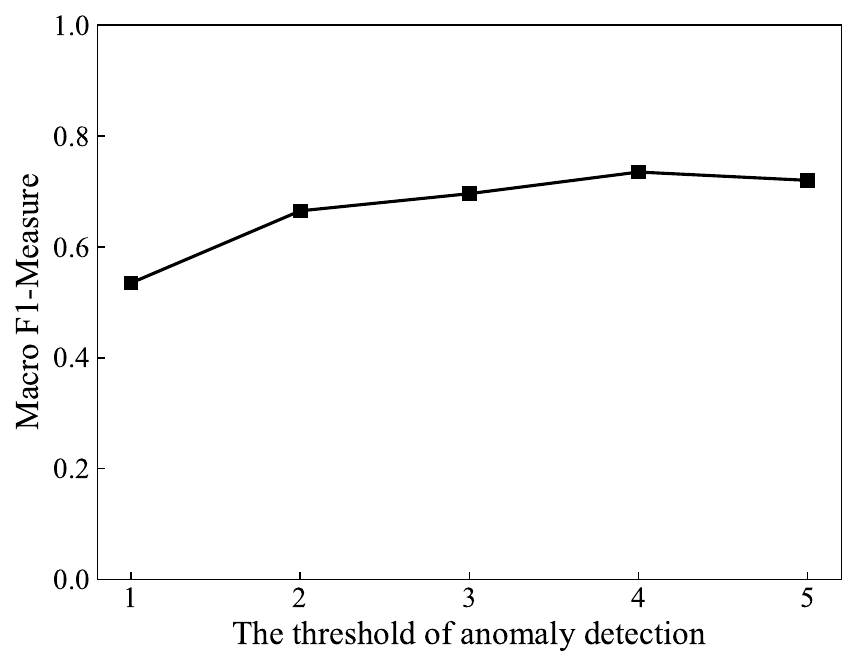}
            \vspace{-10pt}
            \caption{Impact of the threshold of anomaly detection.}
            \label{fig5}
	\end{minipage}%

\end{figure*}

We evaluate the impact of the length of sliding window, increasing it from 9 to 23 with a step size of 2. Fig. \ref{fig3} shows the evaluation results on the HDFS dataset, from which we can observe that with the increase of the length of sliding window, the performance of CSCLog rises gradually. The best performance is achieved when the length of sliding window is 19. The reason is that when the length of sliding window increases from 9 to 19, the patterns of log sequences increase and the ability to capture anomalous information is improved. As the length of sliding window increases from 19 to larger, the performance gradually drops. The reason may be that the increase of sliding window causes the increase of the number of extracted subsequences, which brings a challenge for the model to capture the correlation of subsequences.

We evaluate the impact of the number of top-\textit{k} ranked templates, increasing it from 1 to 11 with a step size of 2. Fig. \ref{fig4} shows the evaluation results on the HDFS dataset, from which we can observe that when the number of top-\textit{k} ranked templates is less than 7, the performance rises as the number increases, which indicates that increasing the number of top-\textit{k} ranked templates within a certain range can improve the fault tolerance of the model. When the number of top-\textit{k} ranked templates is larger than 7, the performance drops slightly as the number increases. The reason may be that when the number is too large, it is easy for the top-\textit{k} ranked templates to hit the true log template, which makes the model misclassifies the anomalous sequences as normal.

We evaluate the impact of the threshold of anomaly detection, increasing it from 1 to 5 with a step size of 1. Fig. \ref{fig5} shows the evaluation results on the HDFS dataset, from which we can observe that with the increase of the threshold of anomaly detection, the performance first increases and then drops. The best performance is achieved when the threshold of anomaly detection is 4. The reason may be that when the threshold is set too small, the log sequences are easy to be detected as anomalies, which contributes to the higher false positive rate and the lower accuracy. When the threshold is set too large, the log sequences are difficult to be detected as anomalies, which contributes to the higher false negative rate and the lower recall.

\subsection{Computation Cost}
We evaluate the computation costs of CSCLog and baselines, including the parameter number, training time, and inference time. Table \ref{tab:cost} shows the evaluation results on the HDFS dataset, from which we can observe that DeepLog has the least parameter number and runs fastest in these methods, but it gets the worst anomaly detection performance. Compared with LogAnomaly, LogC, and OC4Seq, CSCLog gets the best anomaly detection performance while needing the least time for inference, which is more in line with the requirements of modern software systems for real-time anomaly detection. Overall, comprehensively considering the significant anomaly detection performance improvement and the computation costs, CSCLog exhibits its superiority over existing methods.

\begin{table}[!t]
    \caption{The computation costs of different methods. The best results are bold.}
    \resizebox{\linewidth}{!}{
\begin{tabular}{ccccccccc}
\toprule
Model      & \# Parameters   & \begin{tabular}[c]{@{}c@{}}Training time\\ /epoch (s)\end{tabular} & \begin{tabular}[c]{@{}c@{}}Total training \\ time (h)\end{tabular} & \begin{tabular}[c]{@{}c@{}}Inference time\\ /sequence (s)\end{tabular} & \textit{Acc}.           & \textit{Pre}.           & \textit{Rec}.           & \textit{F1}.            \\ \midrule
DeepLog    & \textbf{57,880} & \textbf{172.14}                                                    & \textbf{2.48}                                                      & \textbf{0.0007}                                                        & 0.895          & 0.732          & 0.942          & 0.786          \\ 
LogAnomaly & 428,056         & 258.31                                                             & 4.87                                                               & 0.0016                                                                 & 0.895          & 0.734          & 0.942          & 0.787          \\ 
LogC       & 1,062,944       & 394.45                                                             & 8.76                                                               & 0.0044                                                                 & 0.898          & 0.735          & 0.944          & 0.790          \\ 
OC4Seq     & 402,304         & 284.68                                                             & 4.13                                                               & 0.0014                                                                 & 0.899          & 0.736          & 0.944          & 0.791          \\ 
CSCLog     & 207,207         & 238.55                                                             & 4.02                                                               & 0.0012                                                                 & \textbf{0.966} & \textbf{0.824} & \textbf{0.952} & \textbf{0.895} \\ 
\bottomrule
    \end{tabular}
    }
    \label{tab:cost}
\end{table}


\subsection{Case Study}
To intuitively reveal the superiority of CSCLog, we perform several case studies.

\begin{figure*}[t]
\centering    
    \subfigcapskip=5pt
    \subfigure[DeepLog]{
         \begin{minipage}{0.4\linewidth}
         \centering          
         \includegraphics[width=0.84\linewidth]{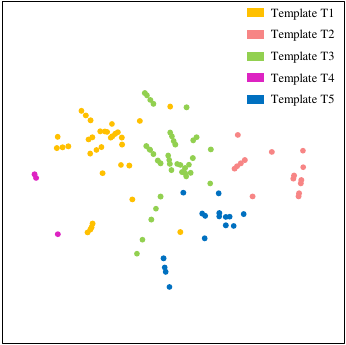}
         \end{minipage}
    }
    \subfigure[CSCLog]{
         \begin{minipage}{0.4\linewidth}
         \centering      
         \includegraphics[width=0.84\linewidth]{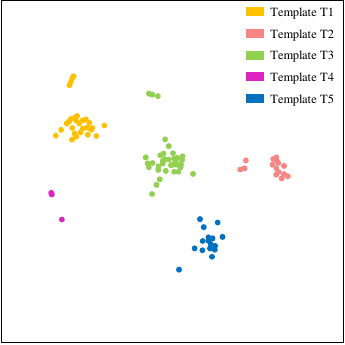}
         \end{minipage}
    }
\caption{Visualization of the embeddings of the output of the final MLP on the OpenStack dataset using t-SNE.} %
\label{fig6}
\vspace{0pt}
\end{figure*}

In Fig. \ref{fig6}, we use the t-SNE method \cite{van2008visualizing} to visualize the embeddings of the output of the final MLP (before the Softmax activation function) on the OpenStack dataset. The log sequence samples whose log templates of the next log message are ``T1'', ``T2'', ``T3'', ``T4'', and ``T5'' are selected. We can find that the sample embeddings of CSCLog are more clustered than those of DeepLog, and the sample embeddings of different log templates are more distantly distributed. It means that CSCLog can better capture the complex log patterns by modeling the implicit correlations of subsequences.

\begin{figure}[!t]
\vspace{10pt}
        \begin{minipage}[t]{1\linewidth}
		\centering
		\includegraphics[width=0.5\linewidth]{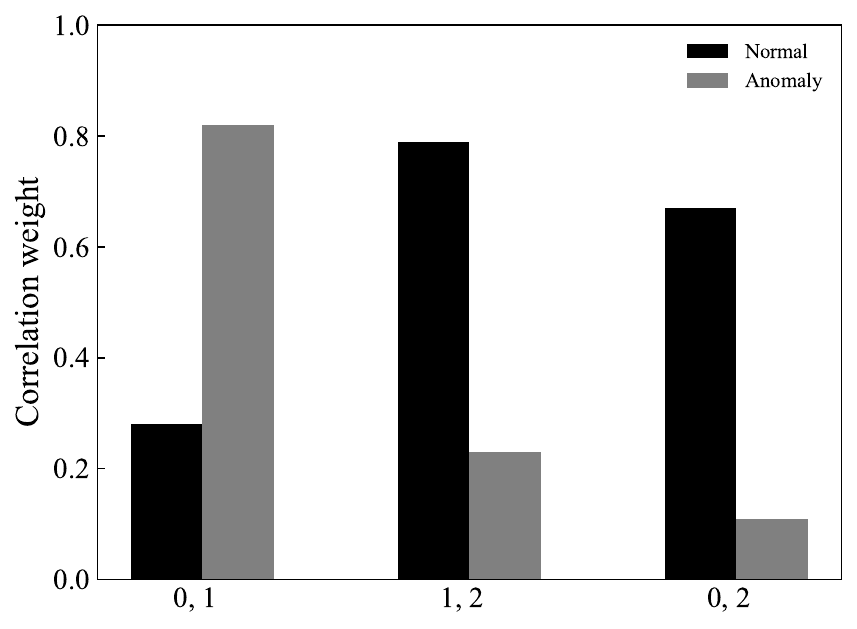}
            \caption{The distribution of the correlation weights.}
            \label{fig7}
	\end{minipage}%
\end{figure}

\begin{table}[!t]
\vspace{15pt}
    \caption{Top-5 ranked templates of two samples predicted by DeepLog and CSCLog. The results are ranked by the predicted probability and the true log templates are bolded.}
    \begin{center}
    \setlength{\tabcolsep}{5.5mm}{
    \begin{tabular}{ccc}
        \toprule
        Model  &     DeepLog	&  CSCLog  \\
        \midrule
        \multirow{5}{*}{Normal}	&Template T3	&\textbf{Template T2}   \\
        	&\textbf{Template T2}	&Template T3   \\
                &Template T5	&Template T6   \\
                &Template T6	&Template T5   \\
                &Template T4	&Template T4   \\
        \cmidrule{1-3}
        \multirow{5}{*}{Anomaly}	&Template T3	&Template T2   \\
        	&Template T2	&Template T3   \\
                &Template T5	&Template T1   \\
                &\textbf{Template T6}	&Template T4   \\
                &Template T4	&Template T5   \\
        \bottomrule
    \end{tabular}}
\label{tab8}
\end{center}
\vspace{-3pt}
\end{table}

Table \ref{tab8} shows the predicted top-5 ranked templates of two log sequence samples on the template prediction task on the OpenStack dataset, where the two samples are labeled as normal and anomaly and their log templates order are both ``T1, T2, T1, T2, T3, T4, T3, T4, T5, T3''. We can find that to the normal sample, both DeepLog and CSCLog predict the log template of the next log message correctly, and CSCLog has a higher ranking to the true log template. To the anomalous sample, the true log template is in the predicted top-5 ranked templates of DeepLog but not in that of CSCLog, which means CSCLog can detect the anomaly while DeepLog cannot. It indicates that CSCLog can improve the ability to capture anomalous information by introducing the implicit correlation encoder and modeling the correlation of subsequences.

Fig. \ref{fig7} shows the distribution of the correlation weights of the subsequences of the above two log sequence samples. These two samples contain components ``nova.scheduler.host.manage'', ``nova.meta-data.wsgi.server'', and ``nova.osapi\_compute.wsgi.server'', whose subsequences are labeled as 0, 1, and 2, respectively. To the normal sample, subsequences 0, 1, and 2 are ``T1, T2, T1, T2'', ``T3, T4, T3'', and ``T3, T4, T5'', respectively. To the anomalous sample, subsequences 0, 1, and 2 are ``T1, T2, T1, T2'', ``T3, T4, T3'', and ``T4, T5, T3'', respectively. We can find that in the normal sample, subsequences 1 and 2 are strongly correlated, and subsequences 0 and 1 are weakly correlated. In contrast, in the anomalous sample, the correlation between subsequences 0 and 1 is extremely strong, and the correlations between the remaining two subsequence pairs are weak. It indicates that CSCLog can capture the complex log patterns of components.

\section{Conclusions and future work}
Anomaly detection based on system logs plays an important role in intelligent operations. In this paper, we propose CSCLog, a Component Subsequence Correlation-Aware Log anomaly detection method, which not only captures the sequential dependencies in subsequences, but also models the implicit correlations of subsequences. Specifically, subsequences are extracted from log sequences based on components and the sequential dependencies in subsequences are captured by LSTMs. The implicit correlation encoder is introduced to model the implicit correlations of subsequences adaptively. In addition, GCNs are employed to accomplish the information interactions of subsequences. We conduct comprehensive experiments on four log datasets and experimental results demonstrate the superiority of CSCLog.

In the future, we will extend this work in the following directions. On the one hand, we will try to introduce hypergraphs or hierarchical graphs to model the higher-order correlations of subsequences, which can improve the representation ability of the model. On the other hand, we will try to introduce the self-attention mechanism with low complexity to improve the parallelism of the model in capturing the sequential dependencies.

\bibliographystyle{ACM-Reference-Format}
\bibliography{arXiv_0704.bib}


\begin{thebibliography}{32}


\ifx \showCODEN    \undefined \def \showCODEN     #1{\unskip}     \fi
\ifx \showDOI      \undefined \def \showDOI       #1{#1}\fi
\ifx \showISBNx    \undefined \def \showISBNx     #1{\unskip}     \fi
\ifx \showISBNxiii \undefined \def \showISBNxiii  #1{\unskip}     \fi
\ifx \showISSN     \undefined \def \showISSN      #1{\unskip}     \fi
\ifx \showLCCN     \undefined \def \showLCCN      #1{\unskip}     \fi
\ifx \shownote     \undefined \def \shownote      #1{#1}          \fi
\ifx \showarticletitle \undefined \def \showarticletitle #1{#1}   \fi
\ifx \showURL      \undefined \def \showURL       {\relax}        \fi
\providecommand\bibfield[2]{#2}
\providecommand\bibinfo[2]{#2}
\providecommand\natexlab[1]{#1}
\providecommand\showeprint[2][]{arXiv:#2}

\bibitem[\protect\citeauthoryear{Bodik, Goldszmidt, Fox, Woodard, and
  Andersen}{Bodik et~al\mbox{.}}{2010}]%
        {bodik2010fingerprinting}
\bibfield{author}{\bibinfo{person}{Peter Bodik}, \bibinfo{person}{Moises
  Goldszmidt}, \bibinfo{person}{Armando Fox}, \bibinfo{person}{Dawn~B Woodard},
  {and} \bibinfo{person}{Hans Andersen}.} \bibinfo{year}{2010}\natexlab{}.
\newblock \showarticletitle{Fingerprinting the datacenter: Automated
  classification of performance crises}. In
  \bibinfo{booktitle}{\emph{Proceedings of the 5th European Conference on
  Computer systems}}. \bibinfo{pages}{111--124}.
\newblock


\bibitem[\protect\citeauthoryear{Chen, Chen, Zhang, Wen, and Yang}{Chen
  et~al\mbox{.}}{2021}]%
        {chen2021multiscale}
\bibfield{author}{\bibinfo{person}{Donghui Chen}, \bibinfo{person}{Ling Chen},
  \bibinfo{person}{Youdong Zhang}, \bibinfo{person}{Bo Wen}, {and}
  \bibinfo{person}{Chenghu Yang}.} \bibinfo{year}{2021}\natexlab{}.
\newblock \showarticletitle{A multiscale interactive recurrent network for
  time-series forecasting}.
\newblock \bibinfo{journal}{\emph{IEEE Transactions on Cybernetics}}
  \bibinfo{volume}{52}, \bibinfo{number}{9} (\bibinfo{year}{2021}),
  \bibinfo{pages}{8793--8803}.
\newblock


\bibitem[\protect\citeauthoryear{Chen, Zheng, Lloyd, Jordan, and Brewer}{Chen
  et~al\mbox{.}}{2004}]%
        {chen2004failure}
\bibfield{author}{\bibinfo{person}{Mike Chen}, \bibinfo{person}{Alice~X Zheng},
  \bibinfo{person}{Jim Lloyd}, \bibinfo{person}{Michael~I Jordan}, {and}
  \bibinfo{person}{Eric Brewer}.} \bibinfo{year}{2004}\natexlab{}.
\newblock \showarticletitle{Failure diagnosis using decision trees}. In
  \bibinfo{booktitle}{\emph{Proceedings of the 1st IEEE International
  Conference on Autonomic Computing}}. \bibinfo{pages}{36--43}.
\newblock


\bibitem[\protect\citeauthoryear{Du, Li, Zheng, and Srikumar}{Du
  et~al\mbox{.}}{2017}]%
        {du2017deeplog}
\bibfield{author}{\bibinfo{person}{Min Du}, \bibinfo{person}{Feifei Li},
  \bibinfo{person}{Guineng Zheng}, {and} \bibinfo{person}{Vivek Srikumar}.}
  \bibinfo{year}{2017}\natexlab{}.
\newblock \showarticletitle{DeepLog: Anomaly detection and diagnosis from
  system logs through deep learning}. In \bibinfo{booktitle}{\emph{Proceedings
  of the 24th ACM Conference on Computer and Communications Security}}.
  \bibinfo{pages}{1285--1298}.
\newblock


\bibitem[\protect\citeauthoryear{Han and Yuan}{Han and Yuan}{2021}]%
        {han2021unsupervised}
\bibfield{author}{\bibinfo{person}{Xiao Han} {and} \bibinfo{person}{Shuhan
  Yuan}.} \bibinfo{year}{2021}\natexlab{}.
\newblock \showarticletitle{Unsupervised cross-system log anomaly detection via
  domain adaptation}. In \bibinfo{booktitle}{\emph{Proceedings of the 30th ACM
  International Conference on Information \& Knowledge Management}}.
  \bibinfo{pages}{3068--3072}.
\newblock


\bibitem[\protect\citeauthoryear{He, Zhu, Zheng, and Lyu}{He
  et~al\mbox{.}}{2017}]%
        {he2017drain}
\bibfield{author}{\bibinfo{person}{Pinjia He}, \bibinfo{person}{Jieming Zhu},
  \bibinfo{person}{Zibin Zheng}, {and} \bibinfo{person}{Michael~R Lyu}.}
  \bibinfo{year}{2017}\natexlab{}.
\newblock \showarticletitle{Drain: An online log parsing approach with fixed
  depth tree}. In \bibinfo{booktitle}{\emph{Proceedings of the 24th IEEE
  International Conference on Web Services}}. \bibinfo{pages}{33--40}.
\newblock


\bibitem[\protect\citeauthoryear{Kingma and Ba}{Kingma and Ba}{2015}]%
        {kingma2015adam}
\bibfield{author}{\bibinfo{person}{Diederik~P Kingma} {and}
  \bibinfo{person}{Jimmy Ba}.} \bibinfo{year}{2015}\natexlab{}.
\newblock \showarticletitle{Adam: A method for stochastic optimization}.
\newblock \bibinfo{journal}{\emph{International Conference on Learning
  Representations}}  \bibinfo{volume}{9} (\bibinfo{year}{2015}),
  \bibinfo{pages}{1--15}.
\newblock


\bibitem[\protect\citeauthoryear{Kipf, Fetaya, Wang, Welling, and Zemel}{Kipf
  et~al\mbox{.}}{2018}]%
        {kipf2018neural}
\bibfield{author}{\bibinfo{person}{Thomas Kipf}, \bibinfo{person}{Ethan
  Fetaya}, \bibinfo{person}{Kuan~Chieh Wang}, \bibinfo{person}{Max Welling},
  {and} \bibinfo{person}{Richard Zemel}.} \bibinfo{year}{2018}\natexlab{}.
\newblock \showarticletitle{Neural relational inference for interacting
  systems}. In \bibinfo{booktitle}{\emph{Proceedings of the 38th International
  Conference on Machine Learning}}. \bibinfo{pages}{2688--2697}.
\newblock


\bibitem[\protect\citeauthoryear{Le and Zhang}{Le and Zhang}{2022}]%
        {le2022log}
\bibfield{author}{\bibinfo{person}{Van~Hoang Le} {and} \bibinfo{person}{Hongyu
  Zhang}.} \bibinfo{year}{2022}\natexlab{}.
\newblock \showarticletitle{Log-based anomaly detection with deep learning: How
  far are we}. In \bibinfo{booktitle}{\emph{Proceedings of the 44th
  International Conference on Software Engineering}}.
  \bibinfo{pages}{1356--1367}.
\newblock


\bibitem[\protect\citeauthoryear{Li, Chen, Jing, He, and Yu}{Li
  et~al\mbox{.}}{2020}]%
        {li2020swisslog}
\bibfield{author}{\bibinfo{person}{Xiaoyun Li}, \bibinfo{person}{Pengfei Chen},
  \bibinfo{person}{Linxiao Jing}, \bibinfo{person}{Zilong He}, {and}
  \bibinfo{person}{Guangba Yu}.} \bibinfo{year}{2020}\natexlab{}.
\newblock \showarticletitle{SwissLog: Robust and unified deep learning based
  log anomaly detection for diverse faults}. In
  \bibinfo{booktitle}{\emph{Proceedings of the 31st IEEE International
  Symposium on Software Reliability Engineering}}. \bibinfo{pages}{92--103}.
\newblock


\bibitem[\protect\citeauthoryear{Liang, Zhang, Xiong, and Sahoo}{Liang
  et~al\mbox{.}}{2007}]%
        {liang2007failure}
\bibfield{author}{\bibinfo{person}{Yinglung Liang}, \bibinfo{person}{Yanyong
  Zhang}, \bibinfo{person}{Hui Xiong}, {and} \bibinfo{person}{Ramendra Sahoo}.}
  \bibinfo{year}{2007}\natexlab{}.
\newblock \showarticletitle{Failure prediction in IBM BlueGene/L event logs}.
  In \bibinfo{booktitle}{\emph{Proceedings of the 7th IEEE International
  Conference on Data Mining}}. \bibinfo{pages}{583--588}.
\newblock


\bibitem[\protect\citeauthoryear{Liao, Lin, Lin, and Tung}{Liao
  et~al\mbox{.}}{2013}]%
        {liao2013intrusion}
\bibfield{author}{\bibinfo{person}{Hung~Jen Liao}, \bibinfo{person}{Chun
  Hung~Richard Lin}, \bibinfo{person}{Ying~Chih Lin}, {and}
  \bibinfo{person}{Kuang~Yuan Tung}.} \bibinfo{year}{2013}\natexlab{}.
\newblock \showarticletitle{Intrusion detection system: A comprehensive
  review}.
\newblock \bibinfo{journal}{\emph{Journal of Network and Computer
  Applications}} \bibinfo{volume}{36}, \bibinfo{number}{1}
  (\bibinfo{year}{2013}), \bibinfo{pages}{16--24}.
\newblock


\bibitem[\protect\citeauthoryear{Liao, Shi, Bai, Wang, and Liu}{Liao
  et~al\mbox{.}}{2017}]%
        {liao2017textboxes}
\bibfield{author}{\bibinfo{person}{Minghui Liao}, \bibinfo{person}{Baoguang
  Shi}, \bibinfo{person}{Xiang Bai}, \bibinfo{person}{Xinggang Wang}, {and}
  \bibinfo{person}{Wenyu Liu}.} \bibinfo{year}{2017}\natexlab{}.
\newblock \showarticletitle{TextBoxes: A fast text detector with a single deep
  neural network}. In \bibinfo{booktitle}{\emph{Proceedings of the 31st AAAI
  Conference on Artificial Intelligence}}. \bibinfo{pages}{4161--4167}.
\newblock


\bibitem[\protect\citeauthoryear{Lin, Zhang, Lou, Zhang, and Chen}{Lin
  et~al\mbox{.}}{2016}]%
        {lin2016log}
\bibfield{author}{\bibinfo{person}{Qingwei Lin}, \bibinfo{person}{Hongyu
  Zhang}, \bibinfo{person}{Jian~Guang Lou}, \bibinfo{person}{Yu Zhang}, {and}
  \bibinfo{person}{Xuewei Chen}.} \bibinfo{year}{2016}\natexlab{}.
\newblock \showarticletitle{Log clustering based problem identification for
  online service systems}. In \bibinfo{booktitle}{\emph{Proceedings of the 38th
  International Conference on Software Engineering Companion}}.
  \bibinfo{pages}{102--111}.
\newblock


\bibitem[\protect\citeauthoryear{Lou, Fu, Yang, Xu, and Li}{Lou
  et~al\mbox{.}}{2010}]%
        {lou2010mining}
\bibfield{author}{\bibinfo{person}{Jian~Guang Lou}, \bibinfo{person}{Qiang Fu},
  \bibinfo{person}{Shengqi Yang}, \bibinfo{person}{Ye Xu}, {and}
  \bibinfo{person}{Jiang Li}.} \bibinfo{year}{2010}\natexlab{}.
\newblock \showarticletitle{Mining invariants from console logs for system
  problem detection}. In \bibinfo{booktitle}{\emph{Proceedings of the 19th
  USENIX Annual Technical Conference}}. \bibinfo{pages}{1--14}.
\newblock


\bibitem[\protect\citeauthoryear{Meng, Liu, Zhu, Zhang, Pei, Liu, Chen, Zhang,
  Tao, Sun, et~al\mbox{.}}{Meng et~al\mbox{.}}{2019}]%
        {meng2019loganomaly}
\bibfield{author}{\bibinfo{person}{Weibin Meng}, \bibinfo{person}{Ying Liu},
  \bibinfo{person}{Yichen Zhu}, \bibinfo{person}{Shenglin Zhang},
  \bibinfo{person}{Dan Pei}, \bibinfo{person}{Yuqing Liu},
  \bibinfo{person}{Yihao Chen}, \bibinfo{person}{Ruizhi Zhang},
  \bibinfo{person}{Shimin Tao}, \bibinfo{person}{Pei Sun}, {et~al\mbox{.}}}
  \bibinfo{year}{2019}\natexlab{}.
\newblock \showarticletitle{LogAnomaly: Unsupervised detection of sequential
  and quantitative anomalies in unstructured logs}. In
  \bibinfo{booktitle}{\emph{Proceedings of the 28th International Joint
  Conference on Artificial Intelligence}}. \bibinfo{pages}{4739--4745}.
\newblock


\bibitem[\protect\citeauthoryear{Nedelkoski, Bogatinovski, Acker, Cardoso, and
  Kao}{Nedelkoski et~al\mbox{.}}{2020}]%
        {nedelkoski2020self}
\bibfield{author}{\bibinfo{person}{Sasho Nedelkoski}, \bibinfo{person}{Jasmin
  Bogatinovski}, \bibinfo{person}{Alexander Acker}, \bibinfo{person}{Jorge
  Cardoso}, {and} \bibinfo{person}{Odej Kao}.} \bibinfo{year}{2020}\natexlab{}.
\newblock \showarticletitle{Self-attentive classification-based anomaly
  detection in unstructured logs}. In \bibinfo{booktitle}{\emph{Proceedings of
  the 20th IEEE International Conference on Data Mining}}.
  \bibinfo{pages}{1196--1201}.
\newblock


\bibitem[\protect\citeauthoryear{Oliner and Stearley}{Oliner and
  Stearley}{2007}]%
        {oliner2007supercomputers}
\bibfield{author}{\bibinfo{person}{Adam Oliner} {and} \bibinfo{person}{Jon
  Stearley}.} \bibinfo{year}{2007}\natexlab{}.
\newblock \showarticletitle{What supercomputers say: A study of five system
  logs}. In \bibinfo{booktitle}{\emph{Proceedings of the 37th Annual IEEE/IFIP
  International Conference on Dependable Systems and Networks}}.
  \bibinfo{pages}{575--584}.
\newblock


\bibitem[\protect\citeauthoryear{Paszke, Gross, Massa, Lerer, Bradbury, Chanan,
  Killeen, Lin, Gimelshein, Antiga, et~al\mbox{.}}{Paszke
  et~al\mbox{.}}{2019}]%
        {paszke2019pytorch}
\bibfield{author}{\bibinfo{person}{Adam Paszke}, \bibinfo{person}{Sam Gross},
  \bibinfo{person}{Francisco Massa}, \bibinfo{person}{Adam Lerer},
  \bibinfo{person}{James Bradbury}, \bibinfo{person}{Gregory Chanan},
  \bibinfo{person}{Trevor Killeen}, \bibinfo{person}{Zeming Lin},
  \bibinfo{person}{Natalia Gimelshein}, \bibinfo{person}{Luca Antiga},
  {et~al\mbox{.}}} \bibinfo{year}{2019}\natexlab{}.
\newblock \showarticletitle{PyTorch: An imperative style, high-performance deep
  learning library}.
\newblock \bibinfo{journal}{\emph{Neural Information Processing Systems}}
  \bibinfo{volume}{32} (\bibinfo{year}{2019}), \bibinfo{pages}{1--12}.
\newblock


\bibitem[\protect\citeauthoryear{Rozemberczki, Scherer, He, Panagopoulos,
  Riedel, Astefanoaei, Kiss, Beres, L{\'o}pez, Collignon,
  et~al\mbox{.}}{Rozemberczki et~al\mbox{.}}{2021}]%
        {rozemberczki2021pytorch}
\bibfield{author}{\bibinfo{person}{Benedek Rozemberczki}, \bibinfo{person}{Paul
  Scherer}, \bibinfo{person}{Yixuan He}, \bibinfo{person}{George Panagopoulos},
  \bibinfo{person}{Alexander Riedel}, \bibinfo{person}{Maria Astefanoaei},
  \bibinfo{person}{Oliver Kiss}, \bibinfo{person}{Ferenc Beres},
  \bibinfo{person}{Guzm{\'a}n L{\'o}pez}, \bibinfo{person}{Nicolas Collignon},
  {et~al\mbox{.}}} \bibinfo{year}{2021}\natexlab{}.
\newblock \showarticletitle{PyTorch geometric temporal: Spatiotemporal signal
  processing with neural machine learning models}. In
  \bibinfo{booktitle}{\emph{Proceedings of the 30th ACM International
  Conference on Information \& Knowledge Management}}.
  \bibinfo{pages}{4564--4573}.
\newblock


\bibitem[\protect\citeauthoryear{Sun, Liu, Wu, Pei, Lin, Ou, and Jiang}{Sun
  et~al\mbox{.}}{2019}]%
        {sun2019bert4rec}
\bibfield{author}{\bibinfo{person}{Fei Sun}, \bibinfo{person}{Jun Liu},
  \bibinfo{person}{Jian Wu}, \bibinfo{person}{Changhua Pei},
  \bibinfo{person}{Xiao Lin}, \bibinfo{person}{Wenwu Ou}, {and}
  \bibinfo{person}{Peng Jiang}.} \bibinfo{year}{2019}\natexlab{}.
\newblock \showarticletitle{BERT4Rec: Sequential recommendation with
  bidirectional encoder representations from transformer}. In
  \bibinfo{booktitle}{\emph{Proceedings of the 28th ACM International
  Conference on Information and Knowledge Management}}.
  \bibinfo{pages}{1441--1450}.
\newblock


\bibitem[\protect\citeauthoryear{Van~der Maaten and Hinton}{Van~der Maaten and
  Hinton}{2008}]%
        {van2008visualizing}
\bibfield{author}{\bibinfo{person}{Laurens Van~der Maaten} {and}
  \bibinfo{person}{Geoffrey Hinton}.} \bibinfo{year}{2008}\natexlab{}.
\newblock \showarticletitle{Visualizing data using t-SNE}.
\newblock \bibinfo{journal}{\emph{Journal of Machine Learning Research}}
  \bibinfo{volume}{9}, \bibinfo{number}{11} (\bibinfo{year}{2008}),
  \bibinfo{pages}{2579--2605}.
\newblock


\bibitem[\protect\citeauthoryear{Wan, Liu, Wang, and Wen}{Wan
  et~al\mbox{.}}{2021}]%
        {wan2021glad}
\bibfield{author}{\bibinfo{person}{Yi Wan}, \bibinfo{person}{Yilin Liu},
  \bibinfo{person}{Dong Wang}, {and} \bibinfo{person}{Yujin Wen}.}
  \bibinfo{year}{2021}\natexlab{}.
\newblock \showarticletitle{GLAD-PAW: Graph-based log anomaly detection by
  position aware weighted graph attention network}. In
  \bibinfo{booktitle}{\emph{Proceedings of the 25th Pacific-Asia Conference on
  Knowledge Discovery and Data Mining}}. \bibinfo{pages}{66--77}.
\newblock


\bibitem[\protect\citeauthoryear{Wang, Chen, Ni, Liu, Chen, and Tang}{Wang
  et~al\mbox{.}}{2021}]%
        {wang2021multi}
\bibfield{author}{\bibinfo{person}{Zhiwei Wang}, \bibinfo{person}{Zhengzhang
  Chen}, \bibinfo{person}{Jingchao Ni}, \bibinfo{person}{Hui Liu},
  \bibinfo{person}{Haifeng Chen}, {and} \bibinfo{person}{Jiliang Tang}.}
  \bibinfo{year}{2021}\natexlab{}.
\newblock \showarticletitle{Multi-scale one-class recurrent neural networks for
  discrete event sequence anomaly detection}. In
  \bibinfo{booktitle}{\emph{Proceedings of the 27th ACM SIGKDD Conference on
  Knowledge Discovery \& Data Mining}}. \bibinfo{pages}{3726--3734}.
\newblock


\bibitem[\protect\citeauthoryear{Wang, Tian, Fang, Chen, and Qin}{Wang
  et~al\mbox{.}}{2022}]%
        {wang2022lightlog}
\bibfield{author}{\bibinfo{person}{Zumin Wang}, \bibinfo{person}{Jiyu Tian},
  \bibinfo{person}{Hui Fang}, \bibinfo{person}{Liming Chen}, {and}
  \bibinfo{person}{Jing Qin}.} \bibinfo{year}{2022}\natexlab{}.
\newblock \showarticletitle{LightLog: A lightweight temporal convolutional
  network for log anomaly detection on the edge}.
\newblock \bibinfo{journal}{\emph{Computer Networks}}  \bibinfo{volume}{203}
  (\bibinfo{year}{2022}), \bibinfo{pages}{108616--108642}.
\newblock


\bibitem[\protect\citeauthoryear{Xia, Bai, Yin, Li, and Xu}{Xia
  et~al\mbox{.}}{2021}]%
        {xia2021loggan}
\bibfield{author}{\bibinfo{person}{Bin Xia}, \bibinfo{person}{Yuxuan Bai},
  \bibinfo{person}{Junjie Yin}, \bibinfo{person}{Yun Li}, {and}
  \bibinfo{person}{Jian Xu}.} \bibinfo{year}{2021}\natexlab{}.
\newblock \showarticletitle{LogGAN: A log-level generative adversarial network
  for anomaly detection using permutation event modeling}.
\newblock \bibinfo{journal}{\emph{Information Systems Frontiers}}
  \bibinfo{volume}{23} (\bibinfo{year}{2021}), \bibinfo{pages}{285--298}.
\newblock


\bibitem[\protect\citeauthoryear{Xu, Huang, Fox, Patterson, and Jordan}{Xu
  et~al\mbox{.}}{2009}]%
        {xu2009detecting}
\bibfield{author}{\bibinfo{person}{Wei Xu}, \bibinfo{person}{Ling Huang},
  \bibinfo{person}{Armando Fox}, \bibinfo{person}{David Patterson}, {and}
  \bibinfo{person}{Michael~I Jordan}.} \bibinfo{year}{2009}\natexlab{}.
\newblock \showarticletitle{Detecting large-scale system problems by mining
  console logs}. In \bibinfo{booktitle}{\emph{Proceedings of the 22nd ACM
  SIGOPS Symposium on Operating Systems Principles}}.
  \bibinfo{pages}{117--132}.
\newblock


\bibitem[\protect\citeauthoryear{Yao, Rosasco, and Caponnetto}{Yao
  et~al\mbox{.}}{2007}]%
        {yao2007early}
\bibfield{author}{\bibinfo{person}{Yuan Yao}, \bibinfo{person}{Lorenzo
  Rosasco}, {and} \bibinfo{person}{Andrea Caponnetto}.}
  \bibinfo{year}{2007}\natexlab{}.
\newblock \showarticletitle{On early stopping in gradient descent learning}.
\newblock \bibinfo{journal}{\emph{Constructive Approximation}}
  \bibinfo{volume}{26}, \bibinfo{number}{2} (\bibinfo{year}{2007}),
  \bibinfo{pages}{289--315}.
\newblock


\bibitem[\protect\citeauthoryear{Yin, Yan, Xu, Xu, Li, Yang, and Zhang}{Yin
  et~al\mbox{.}}{2020}]%
        {yin2020improving}
\bibfield{author}{\bibinfo{person}{Kun Yin}, \bibinfo{person}{Meng Yan},
  \bibinfo{person}{Ling Xu}, \bibinfo{person}{Zhou Xu}, \bibinfo{person}{Zhao
  Li}, \bibinfo{person}{Dan Yang}, {and} \bibinfo{person}{Xiaohong Zhang}.}
  \bibinfo{year}{2020}\natexlab{}.
\newblock \showarticletitle{Improving log-based anomaly detection with
  component-aware analysis}. In \bibinfo{booktitle}{\emph{Proceedings of the
  27th IEEE International Conference on Software Maintenance and Evolution}}.
  \bibinfo{pages}{667--671}.
\newblock


\bibitem[\protect\citeauthoryear{Zhang, Zhang, Moscato, and Zhang}{Zhang
  et~al\mbox{.}}{2020}]%
        {zhang2020anomaly}
\bibfield{author}{\bibinfo{person}{Bo Zhang}, \bibinfo{person}{Hongyu Zhang},
  \bibinfo{person}{Pablo Moscato}, {and} \bibinfo{person}{Aozhong Zhang}.}
  \bibinfo{year}{2020}\natexlab{}.
\newblock \showarticletitle{Anomaly detection via mining numerical workflow
  relations from logs}. In \bibinfo{booktitle}{\emph{Proceedings of the 39th
  IEEE International Symposium on Reliable Distributed Systems}}.
  \bibinfo{pages}{195--204}.
\newblock


\bibitem[\protect\citeauthoryear{Zhang, Li, Zhang, Lu, Hou, Hu, Gui, and
  Lu}{Zhang et~al\mbox{.}}{2021}]%
        {zhang2021logattn}
\bibfield{author}{\bibinfo{person}{Linming Zhang}, \bibinfo{person}{Wenzhong
  Li}, \bibinfo{person}{Zhijie Zhang}, \bibinfo{person}{Qingning Lu},
  \bibinfo{person}{Ce Hou}, \bibinfo{person}{Peng Hu}, \bibinfo{person}{Tong
  Gui}, {and} \bibinfo{person}{Sanglu Lu}.} \bibinfo{year}{2021}\natexlab{}.
\newblock \showarticletitle{LogAttn: Unsupervised log anomaly detection with an
  autoencoder based attention mechanism}. In
  \bibinfo{booktitle}{\emph{Proceedings of the 14th International Conference on
  Knowledge Science, Engineering and Management}}. \bibinfo{pages}{222--235}.
\newblock


\bibitem[\protect\citeauthoryear{Zhang, Xu, Lin, Qiao, Zhang, Dang, Xie, Yang,
  Cheng, Li, et~al\mbox{.}}{Zhang et~al\mbox{.}}{2019}]%
        {zhang2019robust}
\bibfield{author}{\bibinfo{person}{Xu Zhang}, \bibinfo{person}{Yong Xu},
  \bibinfo{person}{Qingwei Lin}, \bibinfo{person}{Bo Qiao},
  \bibinfo{person}{Hongyu Zhang}, \bibinfo{person}{Yingnong Dang},
  \bibinfo{person}{Chunyu Xie}, \bibinfo{person}{Xinsheng Yang},
  \bibinfo{person}{Qian Cheng}, \bibinfo{person}{Ze Li}, {et~al\mbox{.}}}
  \bibinfo{year}{2019}\natexlab{}.
\newblock \showarticletitle{Robust log-based anomaly detection on unstable log
  data}. In \bibinfo{booktitle}{\emph{Proceedings of the 27th ACM Software
  Engineering Conference and Symposium on the Foundations of Software
  Engineering}}. \bibinfo{pages}{807--817}.
\newblock


\end{thebibliography}

\end{document}